\documentclass[10pt,twocolumn,letterpaper]{article}

\usepackage[pagenumbers]{cvpr} 

\usepackage[dvipsnames]{xcolor}

\definecolor{cvprblue}{rgb}{0.21,0.49,0.74}
\usepackage[pagebackref,breaklinks,colorlinks,citecolor=cvprblue]{hyperref}

\usepackage{algorithm}
\usepackage{algorithmic}
\usepackage{multirow}
\usepackage{amssymb}
\usepackage{subcaption}
\usepackage{color}

\usepackage{graphicx}
\usepackage{stfloats}

\def\paperID{14799} 
\def\confName{CVPR}
\def\confYear{2024}

\title{Large Model Based Referring Camouflaged Object Detection}

\author{Shupeng Cheng$^{1}$ ~~~~Ge-Peng Ji$^{2}$ ~~~~Pengda Qin$^{3}$ ~~~~Deng-Ping Fan$^{4}$ ~~~~Bowen Zhou$^{1}$ ~~~~Peng Xu$^{1}$ \\
$^1$Department of Electronic Engineering, Tsinghua University\\
$^2$School of Computing, Australian National University\\
$^3$Alibaba Group~~~~
$^4$Nankai International Advanced Research Institute, Nankai University\\
{\tt\small chengsp20@mails.tsinghua.edu.cn}~~~
{\tt\small gepengai.ji@gmail.com}~~~
{\tt\small pengda.qpd@alibaba-inc.com}\\
{\tt\small dengpfan@gmail.com}~~~~
{\tt\small zhoubowen@tsinghua.edu.cn}~~~
{\tt\small  peng\_xu@tsinghua.edu.cn}
}

\begin{document}
\maketitle

\begin{abstract}
Referring camouflaged object detection (Ref-COD) is a recently-proposed problem aiming to segment out specified camouflaged objects matched with
a textual or visual reference. 
This task involves two major challenges: the COD domain-specific perception and multimodal reference-image alignment.
Our motivation is to make full use of the semantic intelligence and intrinsic knowledge of recent Multimodal Large Language Models (MLLMs) to decompose this complex task in a human-like way.
As language is highly condensed and inductive, linguistic expression is the main media of human knowledge learning, and the transmission of knowledge information follows a multi-level progression from simplicity to complexity.
In this paper, we propose a large-model-based {\bf M}ulti-{\bf L}evel {\bf K}nowledge-{\bf G}uided multimodal method for Ref-COD termed MLKG, 
where multi-level knowledge descriptions from MLLM are organized to guide the large vision model of segmentation
to perceive the camouflage-targets and camouflage-scene progressively and meanwhile deeply align the textual references with camouflaged photos.
To our knowledge, our contributions mainly include:
(1) This is the first time that the MLLM knowledge is studied for Ref-COD and COD.
(2) We, for the first time, propose decomposing Ref-COD into two main perspectives of perceiving the target and scene by integrating MLLM knowledge, and contribute a multi-level knowledge-guided method. 
(3) Our method achieves the state-of-the-art on the Ref-COD benchmark outperforming numerous strong competitors.
Moreover, thanks to the injected rich knowledge, it demonstrates zero-shot generalization ability on uni-modal COD datasets. {\bf We will release our code soon.}
\end{abstract}

\section{Introduction}
\label{sec:intro}

\begin{figure}[!t]
    \centering
    \begin{subfigure}{\linewidth}
        \centering
        \includegraphics[width=\linewidth]{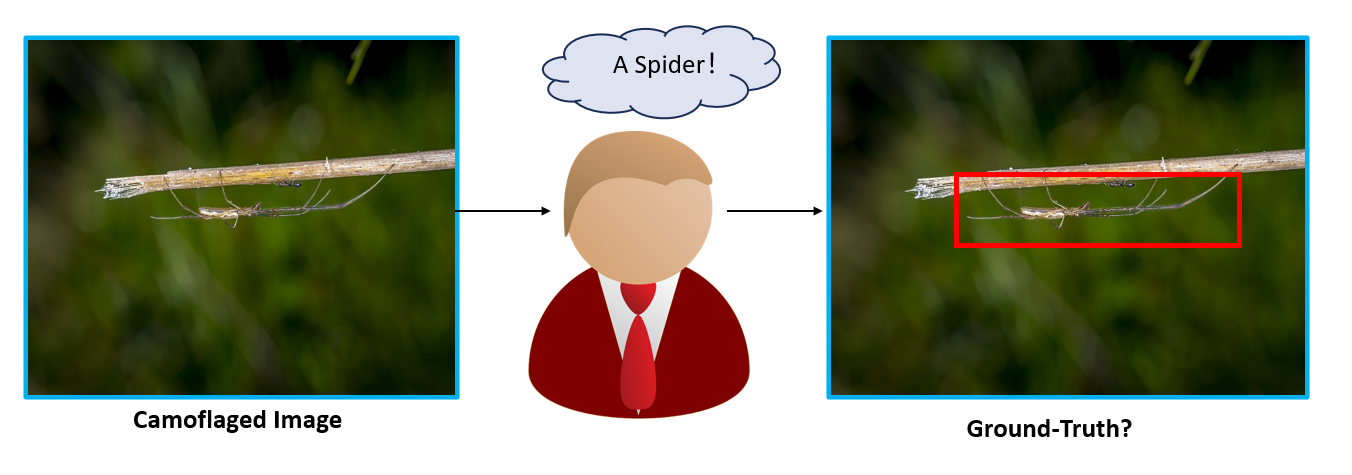}
        \caption{Ignore the truly camouflaged object at first glance}
        \label{fig:sub1}
    \end{subfigure}

    \begin{subfigure}{\linewidth}
        \centering
        \includegraphics[width=\linewidth]{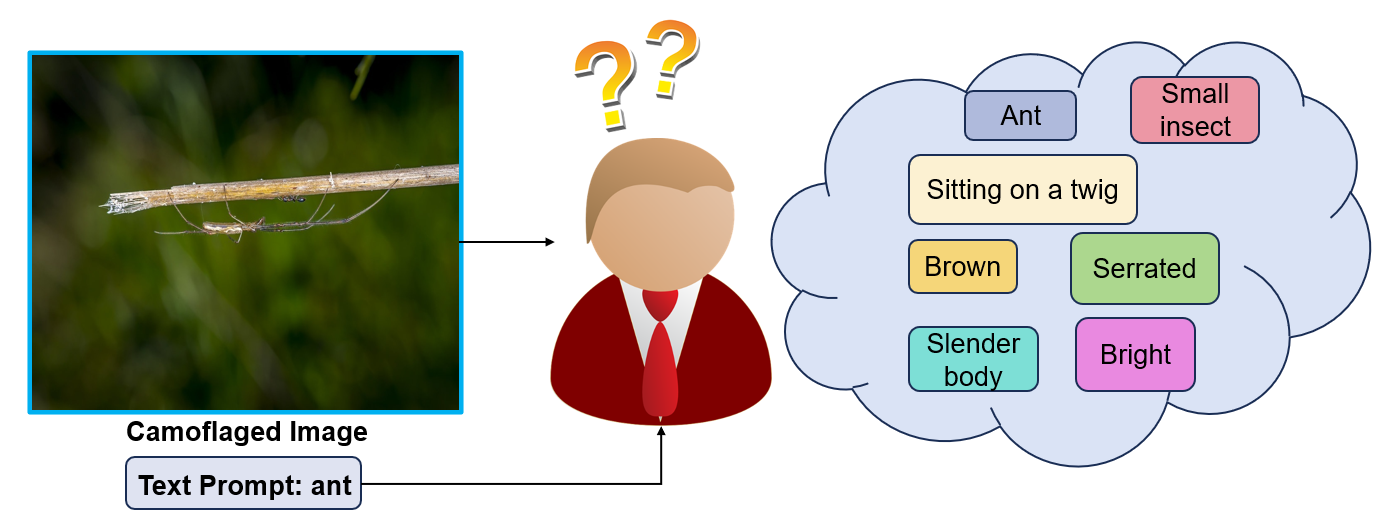}
        \caption{Leverage multi-level knowledge given the text prompt}
        \label{fig:sub2}
    \end{subfigure}

    \begin{subfigure}{\linewidth}
        \centering
        \includegraphics[width=\linewidth]{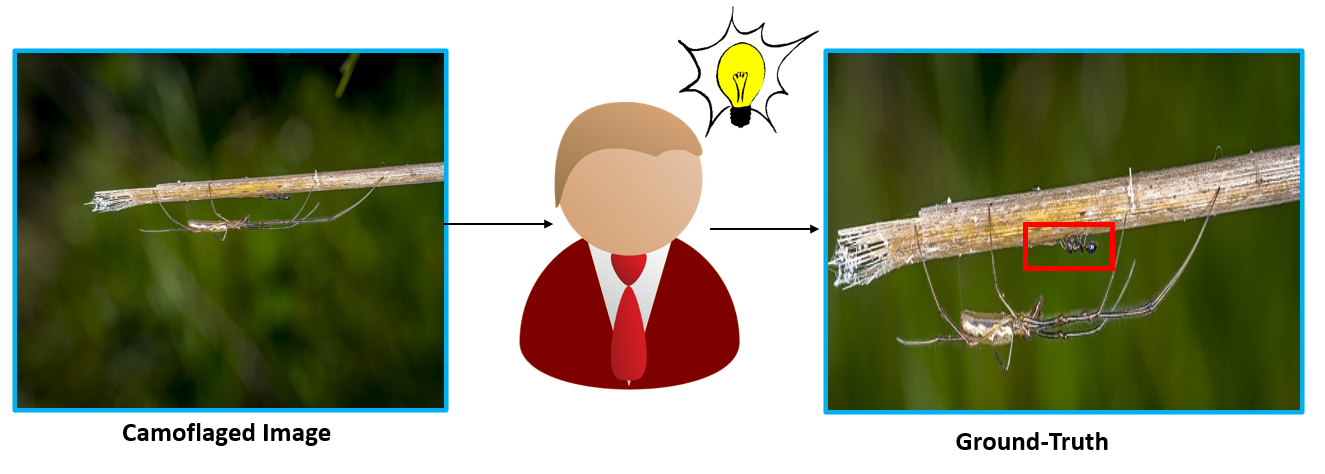}
        \caption{Perceive the camouflaged object under the guidance of multi-level knowledge}
        \label{fig:sub3}
    \end{subfigure}
    \caption{Our motivation: 
    Given a camouflaged photo with a text reference, persons can leverage their inherent multi-level knowledge that is beyond the visual domain to enhance the recognition of subtly camouflaged objects that are prone to be overlooked. Best viewed in colour.}
    \label{fig:entire}
\end{figure}


Camouflaged object detection (COD)~\cite{fan2023advances} is a  vision topic that is widely studied with diverse real-world applications, including rare species discovery, healthcare (\eg, automatic diagnosis of colorectal polyps~\cite{ji2022video}), and agriculture (\eg, pest identification~\cite{liu2019pestnet}).
In 2023, Zhang \etal \cite{zhang2023referring} enhanced this pure vision task further by proposing a multimodal setting, termed referring camouflaged object detection (Ref-COD).
This novel task aims to tackle a kind of practical scenarios where the models segment out specified
camouflaged objects based on a textual or visual reference.
Its goal is to make the process of camouflaged visual perception more orientable and efficient.
In this paper, we focus on the scenarios that use textual reference, \eg,  \textcolor{black}{class label} of the target object.


\subsection{Motivation}
However, Ref-COD is a challenging problem and even the large vision models fail to solve it well \cite{ji2023sam}. We do have to consider at least two major problems.
(1) It is challenging for the models to perceive the complex patterns in the camouflaged scenes.
Even for the human eyes, it is not easy to recognize objects with camouflage properties from a camouflage scene.
As illustrated in Figure \ref{fig:entire},
if a person is presented with a text prompt/reference, that person is likely to combine some prior knowledge other than visual information to solve the Ref-COD problem.
In some sense, this task goes beyond visual perception, thus it is natural for us to consider injecting some domain knowledge into the model to enhance its perception ability. 
(2) The textual references make the models need to handle the cross-modal alignment between the camouflaged photo and its textual reference. If the textual references are too short and abstract to be overly intricate for the models and the models have not seen the relevant corpus during the pre-training, the cross-modal alignment will be more difficult. For instance, if the textual reference is a class label  \texttt{[bird]}, it is easy for the models to associate this concept with the visual attributes of various birds, as  birds are common in the training corpus such that the models have \textcolor{black}{built-in knowledge} of birds \textcolor{black}{and their habitat scenes}. However, let us imagine a difficult case that if
the textual reference is a rare species  \textcolor{black}{\texttt{[platypus]}}, which is infrequent in the training corpus and even the real world. As a result, the models do not have \textcolor{black}{any knowledge} of platypuses \textcolor{black}{and their habitat scenes} to 
align the reference word \textcolor{black}{\texttt{[platypus]}} to the corresponding camouflaged photos of platypuses. 
Considering this limitation, we would use comprehensible/rich natural language to explain and enhance the abstract textual references for Ref-COD. This would mainly have two benefits: 
(a) rich natural language will bring extra knowledge for the models to inject unseen patterns.
(b) and meanwhile make it easier for the models to align the cross-modal inputs.

Nowadays, AI research has entered the era of large models (LM).
New research opportunities are emerging.
Thanks to the multimodal capabilities of Transformers~\cite{xu2023multimodal},
 we see a trend that various LMs (\eg, large language models (LLMs)~\cite{thoppilan2022lamda,lewis2019bart}, multimodal large models~\cite{liu2023llava,liu2023improved}, large vision models~\cite{kirillov2023segment,he2022masked}) have achieved great success on numerous applications.
Given the superior semantic intelligence
and rich knowledge of LMs,
in this paper, we would like to make the first attempt to 
explore LM knowledge
to alleviate the aforementioned task-specific challenges of Ref-COD, and meanwhile implement a solution with interpretability.  

\subsection{Method Overview}
Based on the above motivations and motivated by  \textcolor{black}{LM's fruitful applications on complex task decomposition~\cite{wei2022chain,kojima2022large}},
our idea is to make full use of the semantic intelligence and intrinsic knowledge of MLLMs to decompose this complex task 
in a human-like way.
Specifically,
we would decompose the complex logic of Ref-COD into two main perspectives,  perceiving camouflage-targets and camouflage-scene progressively.  This could guide the model to explicitly figure out two questions. The first question is that \textit{what specific objects it needs to separate}, and the second question is that \textit{what kind of camouflaged scene it handles}. We believe that effectively addressing these two issues can greatly enhance the interpretability of the model.
In this paper, we propose a MLLM-based {\bf M}ulti-{\bf L}evel {\bf K}nowledge-{\bf G}uided multimodal method for Ref-COD termed MLKG, 
where multi-level knowledge descriptions from the popular MLLM LLaVA-1.5~\cite{liu2023improved} are organized to guide the large vision model SAM~\cite{kirillov2023segment}
to perceive the camouflage-target and camouflage-scene progressively and meanwhile deeply align the textual references with camouflaged photos.


\subsection{Contributions}
To our knowledge, our contributions can be mainly summarized as follows.




(1) This is the first work to explore the powerful capabilities of MLLMs to guide the task of Ref-COD and COD, which naturally brings better interpretability.

(2) We, for the first time, propose decomposing Ref-COD into two main perspectives of perceiving the camouflage-target and camouflage-scene by integrating MLLM knowledge to guide the model to explicitly figure out what specific objects it needs to separate from what kind of camouflage scene, and contribute a multi-level knowledge-guided method. 

(3) Our method achieves the state-of-the-art on the Ref-COD benchmark outperforming numerous strong competitors.
Moreover, thanks to the injected rich knowledge, it demonstrates zero-shot generalization ability on uni-modal COD datasets.









\section{Related Work}
\label{sec:related_work}

\subsection{Camouflaged Object Detection}
The field of COD has developed rapidly in recent years. 
As discussed in survey \cite{fan2023advances}, the recent methods can be categorized into three architectures. (1) \textit{ Multi-stream framework} utilizes various feature flows for understanding camouflaged scenes. Those methods are exemplified by generating pseudo-depth maps \cite{wu2023source}, calculating pseudo-edge uncertainty \cite{kajiura2021improving}, employing an adversarial learning paradigm \cite{li2021uncertainty} or a frequency enhancement stream \cite{zhong2022detecting}, and exploiting multi-scale \cite{pang2022zoom} and multi-view \cite{zheng2023mffn} inputs or ensemble backbones \cite{sun2022dqnet}.
(2) \textit{ Bottom-up/top-down framework}, also known as the UNet-shaped framework~\cite{ronneberger2015u}, generates segmentation masks through sequential encoding and decoding procedures. This framework is the most popular in the COD community, with numerous variants such as the method utilizing a coarse-to-fine refinement strategy \cite{sun2021c2fnet} and its enhancement versions \cite{fan2020camouflaged,mei2021camouflaged,fan2021concealed,yin2023camoformer,hu2023high,mei2023distraction,xing2023go,huang2023feature,he2023camouflaged} with a deeply-supervised strategy~\cite{lee2015deeply}.
(3) \textit{ Branched framework} follows a single-input-multiple-output design and is typically formulated as a multi-task learning pipeline. For example, this framework incorporates auxiliary tasks such as confidence estimation~\cite{li2021uncertainty,liu2022modeling,zhang2022preynet,yang2021uncertainty}, target localization/ranking~\cite{lv2021simultaneously,lv2023towards}, category prediction~\cite{le2019anabranch}, depth estimation~\cite{wu2023source,xiang2021exploring}, boundary detection~\cite{zhai2021mutual,zhuge2022cubenet,ji2022fast,zhou2022feature,zhu2022can,sun2022boundary}, and texture exploration~\cite{ji2023gradient,zhu2021inferring}.
In this paper, we focus on the referring camouflaged object detection (Ref-COD) that is a multimodal task. The models aim to segment out specified
camouflaged objects matched with a textual reference.


\subsection{Knowledge Augmented Methods}
The incorporation of high-quality knowledge guidance has been proven to play a crucial role in improving the performance of downstream tasks. Plenty of conventional approaches rely on explicit knowledge injection~\cite{singh2018dock,yu2020cross,cui2021rosita,yu2022survey,mogadala2020integrating}. 
KERL~\cite{chen2018knowledge} introduces a knowledge-embedded image recognition method, where rich visual concepts in the form of knowledge graph are leveraged to improve performance. 
MuRAG~\cite{chen2022murag} resorts to the external knowledge corpus to solve question-answering tasks by retrieving related multimodal knowledge.
As model size continues to increase, large models~\cite{yu2022coca,lewis2019bart,thoppilan2022lamda} exhibit exponential growth in their ability of knowledge capacity and reasoning. Recent Multimodal Large Language Models (MLLMs)~\cite{liu2023llava,ye2023mplug} are able to generate rational analysis when given images and textual instructions. LLaVA-1.5~\cite{liu2023improved} properly aligns large-scale vision encoder with LLM and achieves the state-of-the-art on various multimodal understanding tasks, where the understanding of given images can be expressed as language just like humans.
Based on this, the proposed method treats MLLMs as a high-quality knowledge generator to provide multi-level knowledge for Ref-COD and COD. 

\section{Methodology}
\label{sec:methodology}

\begin{figure*}[!t]
	\centering	
        \includegraphics[width=\linewidth]{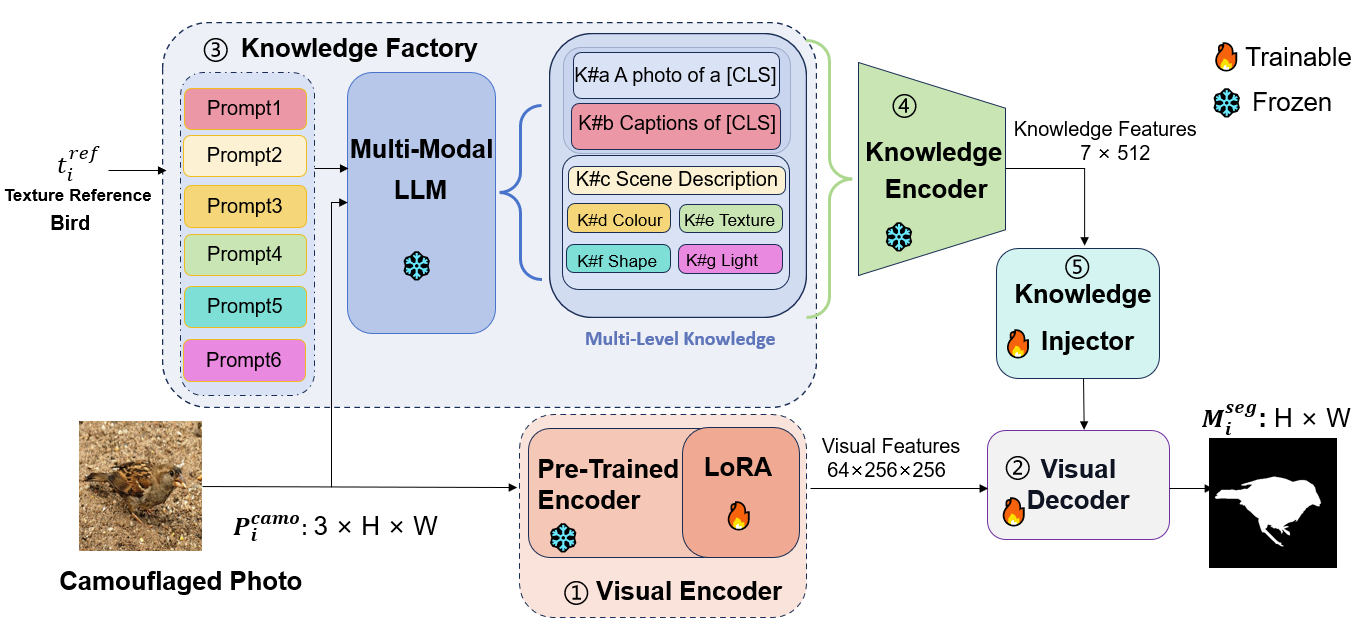}
	\caption{{ Pipeline of our MLKG method. Best viewed in colour.}}
	\label{fig:pipeline}
\end{figure*}

\begin{figure}[!t]
	\centering	
	\includegraphics[width=\linewidth]{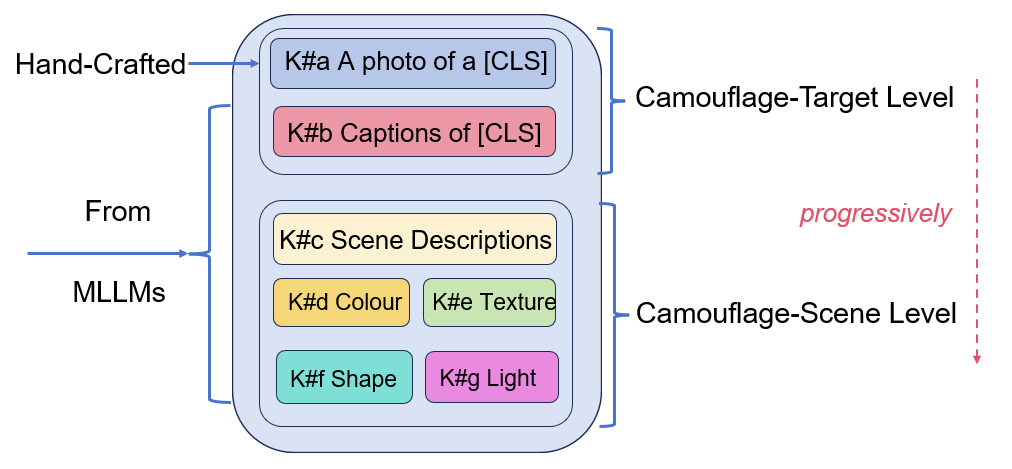}
	\caption{{ Illustration of our multi-level knowledge. Best viewed in colour.}}
	\label{fig:knowledge-sketch}
\end{figure}

\subsection{Formulation}
We assume a training dataset $\mathbf{X}$ in the form of $N$ camouflaged photos:
$\mathbf{X} = \{\mathbf{X}_{i} = (\mathbf{p}^{camo}_{i}, {t}^{ref}_{i})\}^{N}_{i=1}$.
Each camouflaged photo $\mathbf{p}^{camo}_{i} \in \mathbb{R}^{3 \times H \times W}$ has a textual \textcolor{black}{class label}  ${t}^{ref}_{i}$ of the target objects, where $H$ and $W$ denote
the height and width of $\mathbf{p}^{camo}_{i}$.
There may be multiple camouflaged objects in $\mathbf{p}^{camo}_{i}$, and we need to segment out the specific objects by matching the targets provided by the reference ${t}^{ref}_{i}$.
Thus, in Ref-COD task, we aim to learn a multimodal mapping $\mathcal{M}:  \mathbf{X}_{i} \rightarrow \mathbf{M}^{seg}_{i}$
to 
output a binary mask $\mathbf{M}^{seg}_{i} \in \{0, 1\}^{H \times W}$ that can segment out the camouflaged objects from $\mathbf{p}^{camo}_{i}$, which are matching its 
reference ${t}^{ref}_{i}$.

\subsection{Pipeline Overview}
In this work,
we propose a large model based  {\bf m}ulti-{\bf l}evel {\bf k}nowledge-{\bf g}uided multimodal method for Ref-COD, termed MLKG.
As illustrated in Figure~\ref{fig:pipeline}, our pipeline consists of five components:

(1) {\bf Visual Encoder} takes in the camouflaged photo $\mathbf{p}^{camo}_{i}$ and produces visual representation that is transmitted to  the visual decoder;

(2) {\bf Visual Decoder} synchronously receives visual representation and domain knowledge from the visual encoder and knowledge injector respectively, and outputs the segmentation mask $\mathbf{M}^{seg}_{i}$ to predict the specific camouflaged targets matched with the reference ${t}^{ref}_{i}$;

(3) {\bf Knowledge Factory} is a multimodal large language model that translates the target reference ${t}^{ref}_{i}$ as multi-level knowledge descriptions and transmits them to the knowledge encoder;

(4) {\bf Knowledge Encoder} is a text encoder that takes in the multi-level knowledge from the knowledge factory and encodes each level of knowledge into a vector that is passed into the knowledge injector;

(5) {\bf Knowledge Injector} receives the outputs of the knowledge encoder, and completes two signal processing steps:
a)
It selects and fuses the encoded representation of multi-level knowledge received from the knowledge encoder, 
b)
The selected and fused knowledge is input to the visual decoder after cross-modal alignment with the visual decoder's representation space.

Our pipeline is trained in an end-to-end manner with a binary cross entropy (BCE) loss connected to our visual decoder. In particular, we use the configuration of a large vision model SAM~\cite{kirillov2023segment} to implement our visual encoder and decoder. The text encoder of CLIP~\cite{radford2021learning} is adopted as our knowledge encoder. 
In the following sections, we will describe our knowledge factory and injector in detail. See Section~\ref{sec:experiments} for more implementation details of all our components.





\subsection{Knowledge Factory}
In this paper, our textual target reference ${t}^{ref}_{i}$ is a textual class label. We design a multi-level knowledge structure to help the model to decompose Ref-COD task in a human-like manner, perceiving camouflage-targets and camouflage-scene.
This could guide the model to explicitly figure out two questions {\bf Q1}: what specific objects it needs to separate, and {\bf Q2}: what kind of camouflaged scene it handles.
Moreover, we would make full use of the semantic intelligence and intrinsic knowledge of MLLMs to help the model catch up on knowledge related to these two problems.

For Q1, the model requires a clear category name and may need to learn some knowledge of the morphological information or physical characteristics of the specific kind of objects matched with the target reference, disregarding background related interference information. For instance, if the target is a kind of animal, it would not be good for the model to consider the knowledge of their habitat to solve Q1.

For Q2, it would be necessary for the model to learn the knowledge of the camouflaged scene to know which attributes in the scene can cause the camouflage effect of the particular objects, \eg, colour, texture, shape, lighting.

Based on the above analysis, 
we propose a multi-level knowledge guidance (illustrated in Figure~\ref{fig:knowledge-sketch}), \ie, camouflage-target level and camouflage-scene level, which have both hand-crafted texts and detailed language descriptions extracted from MLLMs. 
As listed in Table~\ref{table:prompts}, we designed a series of prompts to 
\textcolor{black}{retrieve} multiple knowledge from MLLMs to organize our domain-specific knowledge.

{\bf Camouflage-Target Level}
has two knowledge components:
{\bf K\#a} (knowledge a): a hand-crafted text template `\texttt{A photo of a [CLS]}' that follows the recently good practice of image classification \cite{radford2021learning} to let the model know what category it needs to perceive. {\bf K\#b}: a detailed description of the morphological information \textcolor{black}{and/or} physical characteristics of the specific target, which is extracted from a 
MLLM by our Prompt 1 (P\#1). 

{\bf Camouflage-Scene Level} provides multiple (from global to fine-grained) knowledge of the camouflaged scene, including five components: 
{\bf K\#c} describes the full scene from a global perspective.
{\bf K\#d}, {\bf K\#e}, {\bf K\#f}, and {\bf K\#g} respectively offer the fine-grained scene description by focusing on colour, texture, shape, and lighting properties.
Knowledge K\#c - K\#g are extracted from a MLLM via using our prompts P\#2 - P\#6, respectively.

After our multi-level knowledge is organized, 
 the knowledge factory module outputs it to the knowledge encoder, as demonstrated in Figure~\ref{fig:pipeline}.
 The knowledge encoder is responsible for transferring knowledge from the linguistic domain to embedding space, via 
 encoding each level of knowledge into a vector that is passed into the knowledge injector.

\begin{table}[!t]
\caption{Our hand-crafted prompts for multimodal LLM. P\#1 is a text prompt, and others are used as multimodal prompts with image inputs. `[CLS]' denotes a placeholder for the class name. 
Double quotes
are used to direct the large model to focus on the key words.}
  \centering
  \resizebox{0.5\textwidth}{!}
   {
  \begin{tabular}{c p{7cm}}
    \toprule
     Num & Content  \\
    \midrule
    P\#1& This is a picture of a [CLS] hidden in the background, please give me a brief description based only on the morphological information of a general [CLS] (Only describe physical characteristics, do not need to describe life habits and so on). \textcolor{black}{No more than 50 words.} \\
    P\#2& Please use detailed language to describe the entire scene in the given image of a [CLS].No more than 30 words. \\        
    P\#3&  Within this scene, how does the [CLS] blend in the environment? Please describe it from the perspective of \textcolor{black}{``colour"} in one sentence. \\
    P\#4& Within this scene, how does the [CLS] blend in the environment? Please describe it from the perspective of ``texture" in one sentence. \\
    P\#5&  Within this scene, how does the [CLS] blend in the environment? Please describe it from the perspective of ``shape" in one sentence. \\
    P\#6& Within this scene, how does the [CLS] blend in the environment? Please describe it from the perspective of ``environment \textcolor{black}{lighting} condition" in one sentence. \\
    \bottomrule
  \end{tabular}
 }
  
  \label{table:prompts}
\end{table}





\subsection{Knowledge Injector}
As the aforementioned,
our knowledge injector receives the outputs of the knowledge encoder, and completes two signal processes:
(a) selecting and fusing the multi-level knowledge and (b) aligning the knowledge with the visual patterns.\\

\begin{figure}[htbp]

	\centering	
	\includegraphics[width=\linewidth]{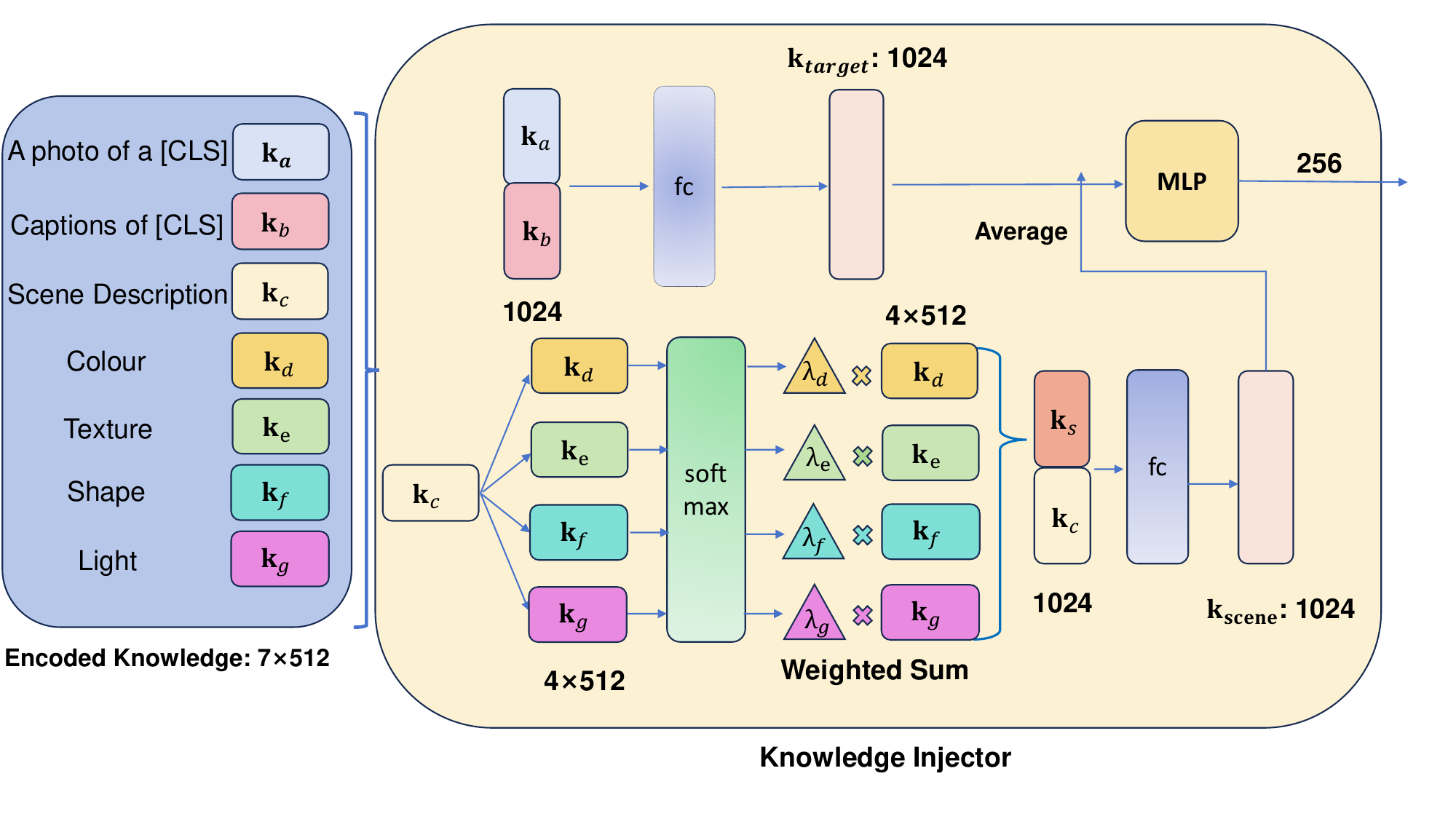}
	\caption{{ Illustration of our knowledge injector. 
 The numbers denote the vector/matrix shapes.
 `fc' means fully connected layer. Best viewed in colour.}}
	\label{fig:knowledge-injector}
\end{figure}

(a)
The knowledge injector selects and fuses the encoded representation of multi-level knowledge received from the knowledge encoder.
For clear interpretability, 
we designed a weighted fusion strategy to select and fuse the multiple encoded knowledge. As demonstrated in Figure~\ref{fig:knowledge-injector}, our knowledge injector adapts the encoded knowledge to inter-level and intra-level selection and fusion, to generate well-integrated guidance for the visual decoder. 
Specifically, knowledge {K\#a} - {K\#g} 
are
encoded into seven 
512d
vectors, denoted as ${\bf k}_a$ - ${\bf k}_g$. 
The encoded vectors ${\bf k}_{a}$ and ${\bf k}_{b}$ of camouflage-target level knowledge { K\#a} and { K\#b} are concatenated as a 
1024d
vector, followed by a 1024 × 1024 fully connected layer, resulting in a 
1024d
vector ${\bf k}_{target}$. For camouflage-scene level knowledge {K\#c} - {K\#g}, we employ a weighted fusion strategy. { K\#c} globally describes the scene of the image, while {K\#d} - {K\#g} separately describes the camouflage scene from the perspectives of colour, texture, shape, and light. 
The proportions of these four aspects vary for different camouflaged photo instances.
Therefore, we use the respective similarity between $\mathbf{k}_c$ and $\mathbf{k}_d$ - ${\bf k}_g$ to determine the knowledge proportions.
We compute the dot product between $\mathbf{k}_c$ and $\mathbf{k}_d$ - ${\bf k}_g$ to obtain four affinity values. After applying the softmax function, these values yield weights $\lambda_d$ - $\lambda_g$. 
By performing a weighted sum on $\mathbf{k}_d - {\bf k}_g$ using $\lambda_d$ - $\lambda_g$, we obtain a new 
512d
vector $\mathbf{k}_s$. Concatenating $\mathbf{k}_s$ and $\mathbf{k}_c$, and passing them through a 1024 × 1024 fully connected layer, we obtain a
1024d
vector $\mathbf{k}_{scene}$. Finally, the average of $\mathbf{k}_{target}$ and $\mathbf{k}_{scene}$ serves as the guidance output by the knowledge injector, which is then injected into the visual decoder. All activation functions for the mentioned fully connected layers above use GELU~\cite{hendrycks2016gaussian}.
(b)
The selected and fused knowledge is input to the visual decoder after cross-modal alignment with the visual decoder's representation space. 
In particular, as stated in our motivation, the injected multi-level knowledge makes it easier for the models to align the cross-modal gaps between the camouflaged photos and knowledge.
Thus, our knowledge injector uses an MLP to project the knowledge into our visual decoder, no need to consider other more complex adapter networks.
Then, the visual decoder synthetically perceives both the visual patterns and knowledge to predict the target masks as the final outputs.

\begin{figure}[!t]

	\centering	
	\includegraphics[width=\linewidth]{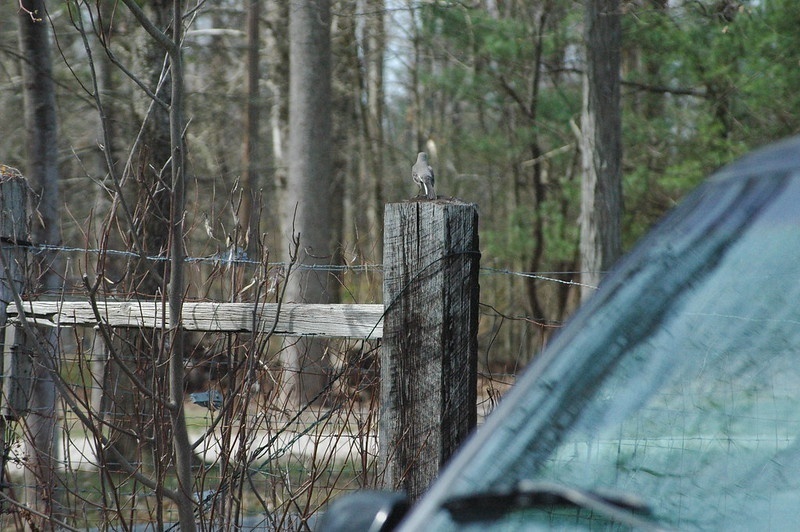}
 	\caption{{A randomly-selected camouflaged photo. In the center of the picture, a bird stands on a stake. Best viewed in colour.}}
 	\label{fig:bird}
 \end{figure}

For instance, we randomly select an example from the class Mockingbird in our testing set, and extract multi-level knowledge from MLLM. See Figure~\ref{fig:bird} and Table~\ref{table:answer}.
In this example, the weights corresponding to `colour', `texture', `shape', and `light' are 0.30, 0.25, 0.27, and 0.18, respectively. This implies that the mockingbird in the image primarily relies on colour to achieve camouflage, because it closely matches the environment. This aligns well with the intuitive perception.

\noindent Here, we would state a minor contribution of ours.
As stated earlier,
our visual encoder and decoder are implemented by a SAM.
However, SAM decoder natively is aligned with the short texts on the category level and its source code for this vision-text alignment has never been released.
In our work, we reproduced this part of the code and
deeply aligned SAM with our complex and structured texts/knowledge.
Compared with the shallow alignment of the original SAM, ours is a more deeply aligning practice. 
This takes the cross-modal alignment for SAM a step forward. 
To our knowledge, our work contributes the first SAM-based Ref-COD solution.


 


\section{Experiments}
\label{sec:experiments}

\subsection{Datasets}
R2C7K~\cite{zhang2023referring} is a recently proposed large-scale dataset for referring camouflaged object detection task. It consists of 2987 camouflaged images for training and 1987 images for testing. In COD field, there are four popular datasets: a) CAMO~\cite{le2019anabranch} has 1000 camouflaged images for training and 250 for testing. b) CHAMELEON~\cite{fan2020camouflaged} consists of  76 images obtained through a Google search for `camouflaged animals' and is solely used for testing purposes. c) NC4K~\cite{lv2021simultaneously}  is currently the largest dataset used to test COD models and consists of 4121 camouflaged images in natural and artificial scenes. d) COD10K~\cite{fan2021concealed} includes 5066 camouflaged iamges, with 3040 for training and 2026 for testing. 
In our experiment, we use  R2C7K~\cite{zhang2023referring} and COD10K~\cite{fan2021concealed} to perform Ref-COD training and testing. 
Furthermore, we evaluate the zero-shot generalization ability of our method on CHAMELEON, NC4K, and the testing set of CAMO.

\subsection{Protocols and Metrics}
Following previous standard protocols in COD, we adopt four widely used metrics to evaluate our method, \ie, structure-measure ($\textit{S}_{m}$)~\cite{fan2017structure}, adaptive E-measure ($\alpha E$)~\cite{fan2018enhanced}, weighted F-measure ($\textit{F}^{w}_{\beta}$)~\cite{margolin2014evaluate}, and mean absolute error ($M$)~\cite{perazzi2012saliency}. Specifically,  $\textit{S}_{m}$~\cite{fan2017structure} measures both object-aware similarity and region-aware  similarity; $\alpha E$~\cite{fan2018enhanced} considers both local and global similarity between two prediction masks; $\textit{F}^{w}_{\beta}$~\cite{margolin2014evaluate} is an exhaustive measure of both recall and precision; $M$~\cite{perazzi2012saliency} is the absolute difference between the prediction mask and the ground-truth.

\begin{table}[!t]
\caption{Our MLLM based multi-level knowledge for
the randomly-selected example in Figure \ref{fig:bird}.}
  \centering
  \resizebox{0.5\textwidth}{!}
   {
  \begin{tabular}{c p{11cm}}
    \toprule
     Knowledge Category & Content  \\
    \midrule
    K\#a & `A photo of a Mockingbird.' \\
    K\#b& `A small bird with a gray body, white underbelly, and black wings. It is perched on a fence post.' \\        
    K\#c&  `A mockingbird perched on a fence post.' \\
    K\#d& `The mockingbird blends in with the environment by perching on a wooden fence post, which is a part of the natural surroundings.' \\
    K\#e&  `The mockingbird's brown and white feathers blend in with the surrounding trees and fence, creating a natural texture that makes it difficult to distinguish from its surroundings.' \\
    K\#f& `The mockingbird blends in the environment by perching on a wooden fence post, which is shaped like a T, and is surrounded by trees and a car.' \\
    K\#g & `The lighting condition in the scene further enhances the bird's blending in, as the sunlight filtering through the trees casts a warm, natural glow on the bird and its surroundings.'\\
    \bottomrule
  \end{tabular}
 }
  
  \label{table:answer}
\end{table}

\begin{table*}[!t]

\caption{Quantitative comparison on Ref-COD benchmark R2C7K~\cite{zhang2023referring} against COD methods and their Ref-COD versions.
`RefT' means `with textual reference'.
`Single-Obj' denotes the scene of a single camouflaged object; `Multi-Obj' denotes the scene of multiple camouflaged objects; `Overall' denotes all scenes that contain camouflaged objects. `$\uparrow/\downarrow$' denotes that the higher/lower the score, the better. The $1^
{st}/2^{
nd}$ best results on column basis are indicated in \textcolor{red}{red}/\textcolor{blue}{blue}.}
  \centering

  \resizebox{0.9\textwidth}{!}
   {
  \begin{tabular}{l*{12}{c}}
    \toprule
    \multirow{2}*{Methods} &\multicolumn{4}{c}{Overall} &\multicolumn{4}{c}{Single-Obj} &\multicolumn{4}{c}{Multi-Obj}\\
    \cmidrule (lr){2-5} \cmidrule (lr){6-9} \cmidrule (lr){10-13}
    &$\textit{S}_{m}\uparrow$ & $\alpha E\uparrow$ & $\textit{F}^{w}_{\beta}\uparrow$ & $\textit{M}\downarrow$   &$\textit{S}_{m}\uparrow$ & $\alpha E\uparrow$ & $\textit{F}^{w}_{\beta}\uparrow$ & $\textit{M}\downarrow$   &$\textit{S}_{m}\uparrow$ & $\alpha E\uparrow$ & $\textit{F}^{w}_{\beta}\uparrow$ & $\textit{M}\downarrow$ \\
    \midrule
    R2CNet-baseline~\cite{zhang2023referring}  &0.772 &0.847 &0.604 &0.044 &0.777 &0.847 &0.611 &0.043 &0.711 &0.849 &0.531 &0.054 \\
    R2CNet-RefT~\cite{zhang2023referring} & 0.806 &0.878 &0.668 &0.037 &0.810  &0.880 &0.674  &0.035  &0.753 &0.870 &0.607  &0.046 \\
    PFNet~\cite{mei2021camouflaged} & 0.791 &0.876 &0.651  &0.040 &0.795  &0.876  &0.656  &0.039  &0.740  &0.868  &0.594 &0.051 \\
    PFNet-RefT~\cite{zhang2023referring}~\cite{mei2021camouflaged}  &0.813  &0.893 &0.693  &0.034  &0.817  &0.892  &0.697  &0.033  &0.769  &0.889 &0.648  &0.041\\
    PreyNet~\cite{zhang2022preynet} &0.806  &0.890 &0.690 &0.034  &0.811  &0.892  &0.696  &0.033  &0.749  &0.878  &0.618  &0.042 \\
    PreyNet-RefT~\cite{zhang2023referring}~\cite{zhang2022preynet}  &0.816  &0.901  &0.705  &0.033  &0.821  &0.900  &0.710  &0.032  &0.759  &\textcolor{blue}{0.902}  &0.648  &0.041 \\
    SINetV2~\cite{fan2021concealed} &0.813 &0.874 &0.678 &0.036 &0.818  &0.874  &0.684  &0.035  &0.763  &0.864  &0.615 &0.045  \\
    SINetV2-RefT~\cite{zhang2023referring}~\cite{fan2021concealed} &0.822 &0.887  &0.696  &0.033  &0.827  &0.888  &0.702  &0.032  &0.766  &0.866  &0.629  &0.043 \\
    BSANet~\cite{zhu2022can} &0.818 &0.893 &0.702 &0.034 &0.823  &0.895  &0.707  &0.033 &0.766 &0.873  &0.643  &0.041\\
    BSANet-RefT~\cite{zhang2023referring}~\cite{zhu2022can}  &0.830  &\textcolor{blue}{0.914}  &0.730  &0.030  &0.834  &\textcolor{blue}{0.915}  &0.734  &0.029  &0.784  &0.898  &{0.674}  &\textcolor{blue}{0.036} \\
    BGNet~\cite{sun2022boundary}  &0.818  &0.901  &0.679  &0.036 &0.822  &0.901  &0.683  &0.035  &0.775  &0.886 &0.626  &0.044 \\
    BGNet-RefT~\cite{zhang2023referring}~\cite{sun2022boundary}  &\textcolor{blue}{0.840}  &0.912  &\textcolor{blue}{0.739}  &\textcolor{blue}{0.029}  &\textcolor{blue}{0.844}  &0.914 &\textcolor{blue}{0.745}  &\textcolor{blue}{0.028}  &\textcolor{blue}{0.791}  &0.888  &\textcolor{blue}{0.677}  &0.038 \\
    ZoomNet~\cite{pang2022zoom} &0.813 &0.884 &0.688 &0.032 &0.818  &0.885  &0.695  &0.031  &0.747  &0.870  &0.605  &0.042 \\
    ZoomNet-RefT~\cite{zhang2023referring}~\cite{pang2022zoom}  &0.835  &0.897  &0.725  &\textcolor{blue}{0.029}  &0.839  &0.897  &0.731  &\textcolor{blue}{0.028}  &0.783  &0.889  &0.661  &0.038 \\
    DGNet~\cite{ji2023deep}  &0.816 &0.883 &0.684 &0.034 &0.826  &0.885  &0.700  &0.032  &0.744 &0.873  &0.588 &0.047 \\
    DGNet-RefT~\cite{zhang2023referring}~\cite{ji2023deep} &0.824  &0.891  &0.701 &0.032  &0.830 &0.892  &0.709  &0.031  &0.745  &0.873  &0.596 &0.046\\
    \midrule
    
    Our MLKG   &\textcolor{red}{0.910} &\textcolor{red}{0.916} &\textcolor{red}{0.829} &\textcolor{red}{0.019} &\textcolor{red}{0.914} &\textcolor{red}{0.917} &\textcolor{red}{0.834} &\textcolor{red}{0.019} &\textcolor{red}{0.856} &\textcolor{red}{0.903} &\textcolor{red}{0.767} &\textcolor{red}{0.029}  \\
    \bottomrule
  \end{tabular}
 }
  
  \label{table1}
  
\end{table*}

\begin{table}[!t]
  \caption{ Quantitative comparison on COD10K~\cite{fan2021concealed} dataset with the state-of-the-art COD methods. `$\uparrow/\downarrow$' denotes that the higher/lower the score, the better. The $1^
{st}/2^{
nd}$ best results on column basis are indicated in \textcolor{red}{red}/\textcolor{blue}{blue}.}
  \centering
 
  \resizebox{0.4\textwidth}{!}
   {
  \begin{tabular}{l*{4}{c}}
    \toprule
    Method &$\textit{S}_{m}\uparrow$ & $\alpha E\uparrow$ & $\textit{F}^{w}_{\beta}\uparrow$ & $\textit{M}\downarrow$  \\
    \midrule
    SINet~\cite{fan2020camouflaged} &0.776 &0.867 &0.631 &0.043  \\
    C2FNet~\cite{sun2021context} &0.813  &0.886 &0.686  &0.036 \\
    PFNet~\cite{mei2021camouflaged} &0.800 &0.868  &0.660  &0.040  \\
    UGTR~\cite{yang2021uncertainty} &0.818   &0.850  &0.667  &0.035  \\
    C2FNet-V2~\cite{chen2022camouflaged} &0.811 &0.890  &0.691  &0.036 \\
    TPRNet~\cite{zhang2022tprnet}   &0.817  &0.869  &0.683  &0.036  \\
    FAPNet~\cite{zhou2022feature} &0.822 &0.875 &0.694 &0.036\\
    BSANet~\cite{zhu2022can}   &0.818  &0.894  &0.699  &0.034 \\
    BGNet~\cite{sun2022boundary} &0.831 &0.902 &0.722 &0.033 \\
    FDNet~\cite{zhong2022detecting} &0.840 &0.906 &0.729 &0.030 \\
    SINetV2~\cite{fan2021concealed}  &0.815 &0.864 &0.680 &0.037 \\
    PopNet~\cite{wu2023source} &0.851 &0.910 &0.757 &0.028 \\
    CRNet~\cite{he2023weakly} &0.733 &0.845 &0.576 &0.049 \\
    PFNet+~\cite{mei2023distraction}  &0.806  &0.880  &0.677  &0.037  \\
    DGnet~\cite{ji2023deep} &0.822 &0.879 &0.693 &0.033 \\
    DTINet~\cite{liu2022boosting}  &0.824  &0.881  &0.695  &0.034  \\
    CamoFormer~\cite{yin2022camoformer}  &0.869  &\textcolor{blue}{0.931}  &0.786  &\textcolor{blue}{0.023} \\ 
    HitNet~\cite{hu2023high}   &\textcolor{blue}{0.871} &\textcolor{red}{0.936} &\textcolor{blue}{0.806}&\textcolor{blue}{0.023}  \\
    \midrule
    Our MLKG &\textcolor{red}{0.910} &0.916 &\textcolor{red}{0.829} &\textcolor{red}{0.019}\\
    \bottomrule
  \end{tabular}
 }
  
  \label{table2}
\end{table}

\begin{table*}[!t]
  \caption{Our zero-shot performance against the fully-training results of the competitors on three COD datasets NC4K~\cite{lv2021simultaneously}, CAMO~\cite{le2019anabranch}, and CHAMELEON~\cite{fan2021concealed}. `$\uparrow/\downarrow$' denotes that the higher/lower the score, the better. The $1^
{st}/2^{
nd}$ best results on column basis are indicated in \textcolor{red}{red}/\textcolor{blue}{blue}. }
  \centering
  \resizebox{0.9\textwidth}{!}{
    \begin{tabular}{l*{12}{c}}
      \toprule
      \multirow{2}*{Method} & \multicolumn{4}{c}{NC4K~\cite{lv2021simultaneously}} & \multicolumn{4}{c}{CAMO~\cite{le2019anabranch}}  & \multicolumn{4}{c}{CHAMELEON~\cite{fan2021concealed}} \\
      \cmidrule(lr){2-5} \cmidrule(lr){6-9}  \cmidrule(lr){10-13} 
      & $\textit{S}_{m}\uparrow$ & $\alpha E\uparrow$ & $\textit{F}^{w}_{\beta}\uparrow$ & $\textit{M}\downarrow$ & $\textit{S}_{m}\uparrow$ & $\alpha E\uparrow$ & $\textit{F}^{w}_{\beta}\uparrow$ & $\textit{M}\downarrow$
      & $\textit{S}_{m}\uparrow$ & $\alpha E\uparrow$ & $\textit{F}^{w}_{\beta}\uparrow$ & $\textit{M}\downarrow$\\
      \midrule
      SINet~\cite{fan2020camouflaged}  &0.808  &0.883  &0.723 &0.058 &0.745 &0.825  &0.644 &0.092 &0.872 &0.936 &0.806 &0.034 \\
      C2FNet~\cite{sun2021context} &0.838 &0.901 &0.762 &0.049 &0.796 &0.865 &0.719 &0.080 &0.886 &0.933 &0.825 &0.033 \\
      UR-COD~\cite{kajiura2021improving} &0.844 &0.910 &0.787  &0.045  &0.814  &0.891  &0.758  &0.067 &0.901 &\textcolor{red}{0.960} &\textcolor{blue}{0.862}  &\textcolor{blue}{0.023}  \\
      C2FNet-V2~\cite{chen2022camouflaged} & 0.840 & 0.900 & 0.770 & 0.048 & 0.799 & 0.869 & 0.730 & 0.077 &0.893  &0.947  &0.845  &0.028 \\
      FAPNet~\cite{zhou2022feature} &0.851  &0.903  &0.775  &0.047  &0.815  &0.877  &0.734  &0.076  &0.893  &0.940  &0.825  &0.028  \\
      BGNet~\cite{sun2022boundary}  &0.851  &0.911 &0.788  &0.044    &0.812 &0.876 &0.749 &0.073 &0.901 &0.943 &0.850 &0.027 \\
      SINetV2~\cite{fan2021concealed} & 0.847 & 0.901 & 0.770 & 0.048 & 0.820 & 0.884 & 0.743 & 0.070 &0.888 &0.941  &0.816  &0.030\\
      DGNet~\cite{ji2023deep} & 0.857 & 0.910 & 0.784 & 0.042 &\textcolor{blue} {0.839} & \textcolor{blue}{0.906} & \textcolor{blue}{0.769}  & \textcolor{blue}{0.057} &0.890 &0.938 &0.816 &0.029 \\
      CamoFormer~\cite{yin2022camoformer} & \textcolor{blue}{0.892} & \textcolor{red}{0.941} & \textcolor{red}{0.847} & \textcolor{red}{0.030} &\textcolor{red} {0.872} & \textcolor{red}{0.931} & \textcolor{red}{0.831} & \textcolor{red}{0.046} &\textcolor{blue}{0.910} &\textcolor{blue}{0.956} &0.859  &\textcolor{blue}{0.023}\\
      
      \midrule
      Our MLKG &\textcolor{red}{0.900} &\textcolor{blue}{0.918} &\textcolor{blue}{0.833} &\textcolor{blue}{0.036}  &0.828 &0.877 &0.744 &0.075  &\textcolor{red}{0.935} &0.941 &\textcolor{red}{0.875} & \textcolor{red}{0.020} \\
      \bottomrule
    \end{tabular}
  }
  
  \label{table3}
\end{table*}

\begin{table}[!t]
\caption{
Ablation results on R2C7K~\cite{zhang2023referring} based on the model component and knowledge variants of our MLKG method. 
`$\uparrow/\downarrow$' denotes that the higher/lower the score, the better. 
`VE': Visual Encoder; `VD': Visual Decoder; `KF': Knowledge Factory; `KE': Knowledge Encoder; `KI': Knowledge Injector. `KF\{K\#a\}' denotes `Knowledge Factory produces knowledge K\#a'. `KF\{K\#a-K\#g\}' denotes `Knowledge Factory produces all seven kinds of knowledge K\#a, K\#b, ..., K\#g'.
}
  \centering
  \resizebox{0.5\textwidth}{!}
   {
  \begin{tabular}{l*{4}{c}}
    \toprule
    Configurations &$\textit{S}_{m}\uparrow$ & $\alpha E\uparrow$ & $\textit{F}^{w}_{\beta}\uparrow$ & $\textit{M}\downarrow$  \\
    \midrule
    VE+VD   &0.895 &0.888 &0.788 &0.023 \\
    VE+VD+KF\{K\#a\}+KE+KI  &0.904 &0.916 &0.819 &0.020\\
    VE+VD+KF\{K\#b\}+KE+KI &0.906 &0.914 &0.822 &0.020 \\
    VE+VD+KF\{K\#c-K\#g\}+KE+KI &0.906  &0.914  &0.818  &0.020  \\
    VE+VD+KF\{K\#a-K\#g\}+KE+KI &0.910 &0.916 &0.829 &0.019 \\
    
    \bottomrule
  \end{tabular}
 }
  
  \label{table4}
\end{table}

\subsection{Implementation Details}
We implement our method using the PyTorch~\cite{paszke2019pytorch} library. Specifically, we choose the image encoder and decoder of SAM~\cite{kirillov2023segment} as our visual encoder and decoder, respectively. The parameters are initialized with SAM's public checkpoint ViT-H~\cite{kirillov2023segment}. 
SAM encoder is a heavy network, thus we fine-tune it with LoRA~\cite{hu2021lora} to preserve its powerful representational capacity while reducing computational complexity. 
We choose CLIP text encoder as our knowledge encoder, and its parameters are initialized with CLIP's public checkpoint ViT-B/16~\cite{radford2021learning} and are frozen 
during our training.
SGD with momentum 0.9 is used as our optimizer. Our learning rate is initially set to 5e-3 and decays following the cosine learning rate strategy. All our experiments are conducted on 
an
NVIDIA 
GeForce
RTX 3090 GPU.


\subsection{Result Analysis}
To demonstrate the effectiveness of our method, we compare it with numerous strong competitors in COD field. 
In Tables \ref{table1} and \ref{table2}, we observed that our model 
obviously
outperforms the existing competitors in three key metrics: $\textit{S}_{m}$, $\textit{F}^{w}_{\beta}$, and $M$. \textcolor{black}{Although our method may sometimes not be optimal in the indicator of $\alpha E$,} the overall performance of our model exhibits
obvious
improvements when compared 
with
the existing COD methods on R2C7K and COD10K.

\subsection{Zero-Shot Generalization}
The multi-level knowledge provides our model with extra rich domain knowledge,  thus this helps our models to understand the unseen patterns.  
To assess the zero-shot generalization capability of our model, direct testing was conducted on three uni-modal COD datasets, \ie, NC4K, CAMO, and CHAMELEON, without any adaptation training. We chose nine strong COD models as competitors. 
\textcolor{black}{As reported in Table \ref{table3}, our approach holds its own when compared to the COD methods fully-trained on each dataset, consistently achieving top-tier performance in most scenarios, frequently ranked the first or second position. While our method may not obtain the state-of-the-art on CAMO, it may need to consider the uncertainties caused by the dataset sizes (NC4K with 4121 images, CAMO with 250 images, and CHAMELEON with 76 images). Based on this phenomenon, we 
 draw a conclusion that
our designed multi-level knowledge
empowers our model with robust zero-shot generalization capabilities.}



\subsection{Ablation Study}

We designed a series of ablation experiments on R2C7K,
to evaluate the effectiveness of each key component of our designed multi-level knowledge.
\textcolor{black}{As illustrated in Table \ref{table4}, whether introducing {\bf K\#a}, K\#b, or {\bf K\#c}-K\#g individually, the rich knowledge derived from the MLLM significantly aids our model in achieving the more precise identification.
Compared with K\#a, K\#b obtains better performance (see the second and third rows in Table \ref{table4}),  demonstrating the effectiveness of comprehensible/rich natural language over abstract textual references. 
Furthermore, when simultaneously injecting knowledge from both camouflage-target and camouflaged-scene levels into our model, it 
helps
our model to decompose the complex task of Ref-COD in a human-like way, resulting in enhanced overall performance.
}

\subsection{Visualization}
\begin{figure*}[!t]
	\centering	
      \includegraphics[width=\linewidth, height=0.75\textheight]{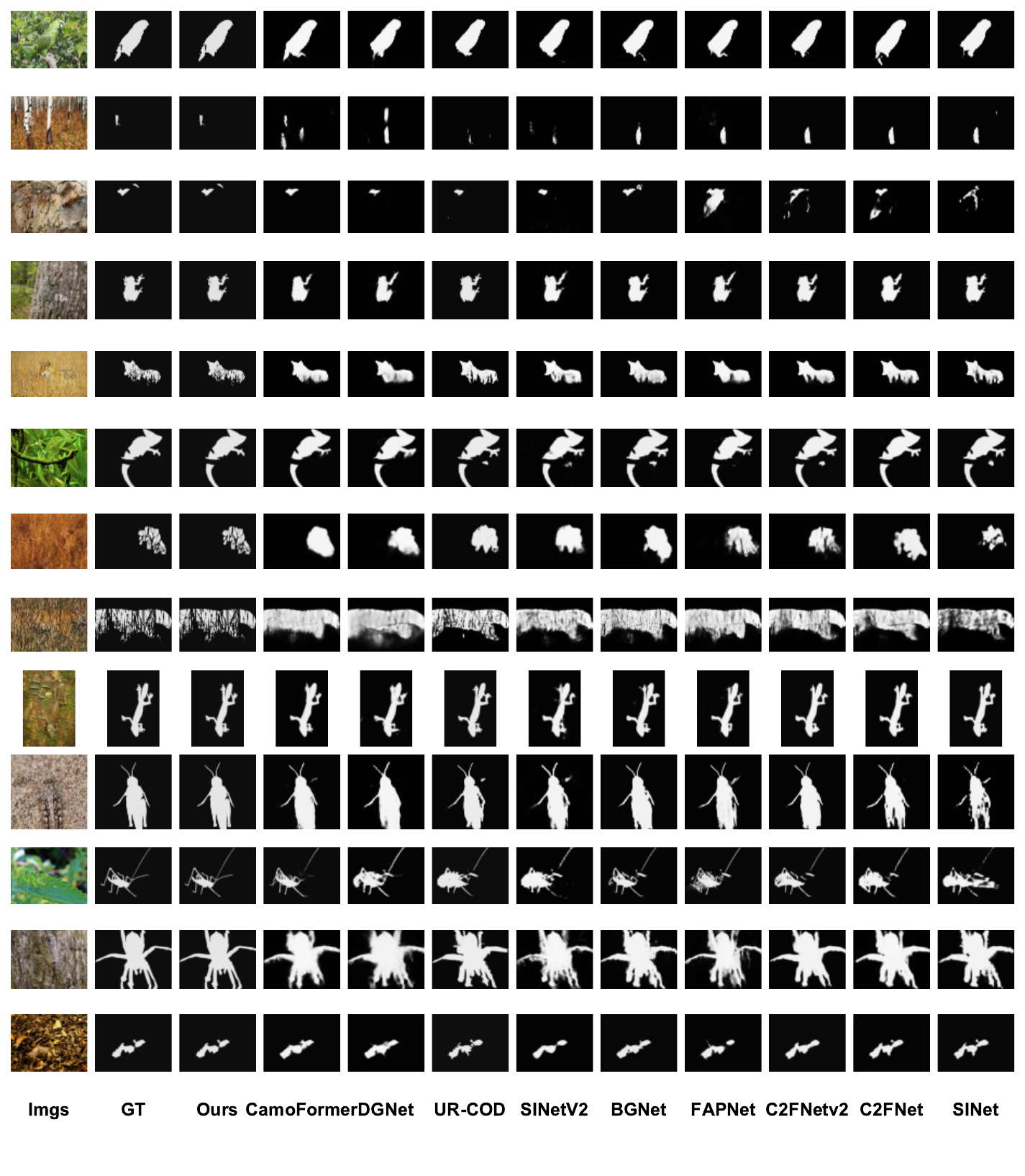}
	\caption{Visual comparison of the proposed method MLKG with nine strong COD methods. 
 The first and second columns are camouflaged photos and their ground truth masks, respectively. Each of the remaining columns arranges the predictions of one method.
 The different shapes are due to the original different resolutions of these photos.
 }
	\label{fig:visual}
\end{figure*}

Figure \ref{fig:visual} presents the qualitative comparison of our model with nine COD models. Compared with these strong competitors, our method 
achieves
more accurate prediction masks.
It is observed that our prediction masks present more subtle contour details of the targets, which are indistinguishable in the camouflaged scenes.
This shows that our model understands the target object and the global scene in a more precise way. 
We would attribute this advantage to its mechanism of complex task decomposition and external knowledge injection.

\section{Conclusion}
\label{sec:conclusion}

This paper proposes a Multi-Level Knowledge-Guided (MLKG) multimodal method for Referring Camouflaged Object Detection (Ref-COD). The method leverages the semantic intelligence and intrinsic knowledge of Multimodal Large Language Models (MLLMs) to decompose the complex task of Ref-COD in a human-like way. The MLKG method organizes multi-level knowledge descriptions from MLLMs to guide the segmentation model in perceiving camouflage-targets and camouflage-scene progressively while aligning textual references with camouflaged photos. The contributions of this paper include exploring the use of MLLMs for Ref-COD and COD, decomposing Ref-COD into two perspectives for better interpretability, and achieving state-of-the-art performance on the Ref-COD and COD benchmark. Additionally, the method demonstrates zero-shot generalization ability on uni-modal COD datasets due to the reasonable injection of rich knowledge.

We hope this work can provide a basis for future work in Ref-COD, COD, and even camouflaged vision perception, motivating
the community toward complex pattern learning using large model knowledge.

\section{Future Work}
\label{sec:future}
In our future efforts, we would try to extend this work mainly in the following three aspects.

(1) Future work can introduce more useful feature information into our current language-based knowledge structure, \eg, latent knowledge from the hidden spaces of some models, structured knowledge from knowledge graphs. Moreover, it is necessary to consider incorporating other modalities, such as depth information and optical information, to further enhance the performance of Ref-COD. This can lead to more comprehensive and robust models that can handle diverse real-world scenarios. 

(2) The rapid development of Large Model brings more and more amazing features. Therefore, there is still room for further exploration of MLLMs for COD tasks. Future work can investigate how newly-updated sophisticated MLLMs can help provide even better performance and interpretability for COD.

(3) The proposed MLKG has been proven to be effective in Ref-COD, but the principle is universally applicable. Future work can not only evaluate the performance of MLKG on more diverse and challenging datasets, but also focus on how to apply it to other visual and multimodal problems. This can help assess the generalization capability of the method and identify potential limitations or areas for improvement.

{
    \small
    \bibliographystyle{ieeenat_fullname}
    \bibliography{main}
}


\end{document}


\setcounter{page}{1}
\maketitlesupplementary

\section{Details about  Knowledge Injector}
\label{sec:rationale}
\begin{figure}[htbp]

	\centering	
	\includegraphics[width=\linewidth]{./figures/Injector.pdf}
	\caption{{ Illustration of our knowledge injector. `fc' means fully connected layer. Best viewed in colour.}}
	\label{fig:knowledge-injector}
\end{figure}

\noindent 
As demonstrated in Figure~\ref{fig:knowledge-injector}, our knowledge injector adapts the encoded knowledge to inter-level and intra-level selection and fusion, to generate well-integrated guidance for the visual decoder. 
Specifically, knowledge {\bf K\#a} - {\bf K\#g} 
are
encoded into seven 
512d
tensors, denoted as $T_a - T_g$. Camouflage-target level knowledge {\bf K\#a} and {\bf K\#b} are concatenated into a 
1024d
tensor, followed by a 1024 × 1024 fully connected layer, resulting in a 
1024d
tensor $T_{target}$. For camouflage-scene level knowledge {\bf K\#c} - {\bf K\#g}, we employ a weighted fusion strategy. {\bf K\#c} globally describes the scene of the image, while {\bf K\#d} - {\bf K\#g} separately describe the camouflage scene from the perspectives of colour, texture, shape, and light. 
The proportions of these four aspects vary for different camouflaged photo instances.
Therefore, we use the similarity between \(T_c\) and \(T_d - T_g\) to determine the knowledge proportions.
We compute the dot product between \(T_c\) and \(T_d - T_g\) to obtain four affinity values. After applying the softmax function, these values yield weights $W_d - W_g$ for the four aspects. By performing a weighted sum on \(T_d - T_g\) using $W_d - W_g$, we obtain a new 
512d
tensor \(T_{s}\). Concatenating \(T_{s}\) and \(T_c\), and after passing through another 1024 × 1024 fully connected layer, we obtain a
1024d
tensor \(T_{scene}\). Finally, the average of \(T_{target}\) and \(T_{scene}\) serves as the guidance output by the knowledge injector, which is then injected into the visual decoder. All activation functions for the mentioned fully connected layers above use GELU.

\section{Visualization Example}
\begin{figure}[!t]

	\centering	
	\includegraphics[width=\linewidth]{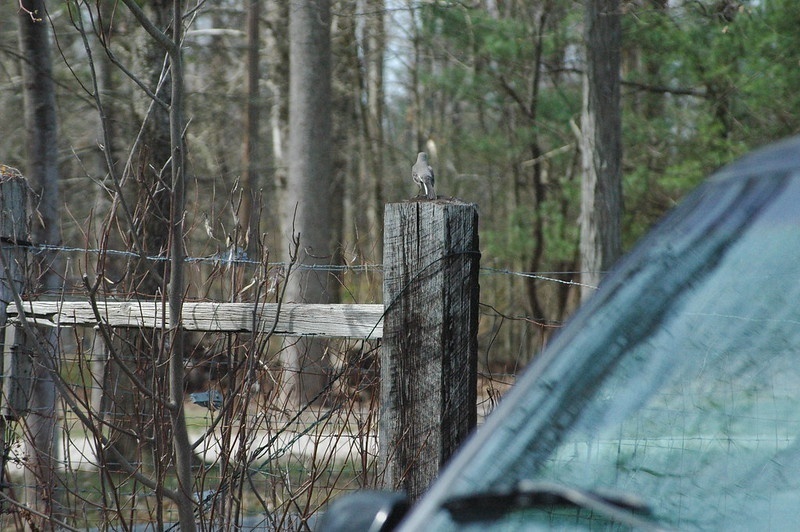}
	\caption{{A randomly-selected camouflaged photo. In the center of the picture, a bird stands on a stake. Best viewed in colour.}}
	\label{fig:bird}
\end{figure}

We randomly select an example from the class Mockingbird in our testing set, and extract multi-level knowledge from MLLM. See Figure~\ref{fig:bird} and Table~\ref{table:answer}.
In this example, the weights corresponding to `colour', `texture', `shape', and `light' are 0.30, 0.25, 0.27, and 0.18, respectively. This implies that the mockingbird in the image primarily relies on colour to achieve camouflage, because it closely matches the environment. This aligns well with the intuitive perception.

\begin{table}[!t]
\caption{Our MLLM based multi-level knowledge for
the randomly-selected example in Figure \ref{fig:bird}.}
  \centering
  \resizebox{0.5\textwidth}{!}
   {
  \begin{tabular}{c p{11cm}}
    \toprule
     Knowledge Category & Content  \\
    \midrule
    K\#a & `A photo of a Mockingbird.' \\
    K\#b& `A small bird with a gray body, white underbelly, and black wings. It is perched on a fence post.' \\        
    K\#c&  `A mockingbird perched on a fence post.' \\
    K\#d& `The mockingbird blends in with the environment by perching on a wooden fence post, which is a part of the natural surroundings.' \\
    K\#e&  `The mockingbird's brown and white feathers blend in with the surrounding trees and fence, creating a natural texture that makes it difficult to distinguish from its surroundings.' \\
    K\#f& `The mockingbird blends in the environment by perching on a wooden fence post, which is shaped like a T, and is surrounded by trees and a car.' \\
    K\#g & `The lighting condition in the scene further enhances the bird's blending in, as the sunlight filtering through the trees casts a warm, natural glow on the bird and its surroundings.'\\
    \bottomrule
  \end{tabular}
 }
  
  \label{table:answer}
\end{table}